\def\year{2022}\relax
\documentclass[letterpaper]{article} % DO NOT CHANGE THIS
\usepackage{aaai22}  % DO NOT CHANGE THIS
\usepackage{times}  % DO NOT CHANGE THIS
\usepackage{helvet}  % DO NOT CHANGE THIS
\usepackage{courier}  % DO NOT CHANGE THIS
\usepackage[hyphens]{url}  % DO NOT CHANGE THIS
\usepackage{graphicx} % DO NOT CHANGE THIS
\usepackage{cite}
\usepackage{amsmath}
\usepackage{amsfonts}
\usepackage{bm}
\usepackage{amssymb}
\usepackage{makecell}
\usepackage{subfigure}
\usepackage{natbib}  % DO NOT CHANGE THIS AND DO NOT ADD ANY OPTIONS TO IT
\usepackage{caption} % DO NOT CHANGE THIS AND DO NOT ADD ANY OPTIONS TO IT
\DeclareCaptionStyle{ruled}{labelfont=normalfont,labelsep=colon,strut=off} % DO NOT CHANGE THIS
\usepackage{algorithm}
\usepackage{algorithmic}

\usepackage{newfloat}
\usepackage{listings}
\floatstyle{ruled}
\newfloat{listing}{tb}{lst}{}
\floatname{listing}{Listing}
\title{BA-Net: Bridge Attention for Deep Convolutional Neural Networks}
\author{
    Yue Zhao\textsuperscript{\rm 1}, Junzhou Chen\textsuperscript{\rm 2},Zirui Zhang\textsuperscript{\rm 3} and Ronghui Zhang\textsuperscript{\rm 4}\\
    School Of Intelligent Systems Engineering\\
    SUN YAT-SUN University
}
\affiliations {
    zhaoy376@mail2.sysu.edu.cn
}
\usepackage{bibentry}
\begin{document}

\maketitle

\begin{abstract}
In attention mechanism research, most existing methods are hard to balance utilizing well the information of the neural network and high computing efficiency, due to heavy feature compression in the attention layer. To solve this problem, this paper proposes a simple and general approach, named Bridge Atttention Net. As a new idea, BA-Net straightforwardly integrates features from previous layers and effectively promotes information interchange. To find the best BA-Net structure, we extensively investigate the effectiveness of previous features of different types and distances and evaluate different configurations of BA structures. Through extensive experimental evaluation, we discovered a simple and exciting insight. That is, bridging all the convolution outputs inside each block with BN can obtain better attention to enhance the performance of the existing neural network architectures. It is effective, stable and easy to use. Moreover, we have released the source code on https://github.com/zhaoy376/Bridge-Attention.
\end{abstract}

\section{Introduction}
Deep convolutional neural networks (CNNs) are widely used in the computer vision community\cite{xu2015show}, showing excellent performance on various tasks, e.g., image classification, object detection, instance segmentation, and semantic segmentation. Since the appearance of AlexNet \cite{krizhevsky2012imagenet}, there have been numerous research dedicated to boosting the performance of CNNs \cite{simonyan2014very,he2016deep,huang2017densely,wang2018non}. 

In recent years, attention mechanism, as a novel technique to enhance performance, has attracted much attention. It learns attention weights from the adjacent convolution layer, thus concentrating on more important features. The channel attention mechanism is one of the attention mechanisms with the most representative method such as squeeze-and-excitation networks (SENet) \cite{hu2018squeeze}, which learns the channel attention from an average pooled output on each map, bringing considerable performance gain in various CNNs. %Another part is spatial attention mechanism, like CBAM \cite{woo2018cbam} and BAM \cite{park2018bam}, learning spatial attention weights to act on the different areas on the feature maps. 

\begin{figure} 
\centering
\includegraphics[height=3cm, width=7.4cm]{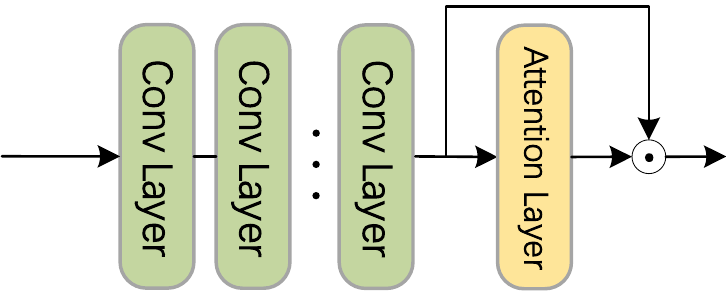} \\
\caption{The block structure of most existing attention methods. The output, coming from stacked convolution layers, passes through the attention layer,thus the attention weights is obtained. }\label{Most Attention Module} 
\end{figure}

Fig.\ref{Most Attention Module} shows the block architecture of most attention methods, which consists of stacked convolution layers and one attention layer. The attention layer utilizes the output of the adjacent convolution layer, which is always the last convolution layer. To enhance the attention weights, many attention methods \cite{bello2019attention,gao2019global} attempt to extract various features from the adjacent convolution layer for information interchange, however, increase the model complexity and computation cost. Some recent attention mechanisms \cite{wang2020eca,zhang2021sa} use more efficient approaches with 1D convolution and channel shuffling, respectively. As a result, most researches are hard to balance utilizing well the information of the neural network and high computing efficiency, due to heavy feature compression of the input features. Learning attention weights from adjacent layer seems to have reached a bottleneck.

Our method provide a new thinking, straightforwardly integrates features from previous layers and effectively promotes information interchange. Moreover, we extensively investigate the effectiveness of previous features of different types and distances and evaluate different configurations of BA structures. As a conclusion, bridging all the convolution outputs inside each block can obtain better attention to enhance the performance of the existing neural network architectures. We implement the BA module only by full connection(FC) and batch normalization(BN). Without complicated strategies, richer information is expected to be directly absorbed by the attention layer.

We implement the BA-Net in simple way based on SENet, thus the BA-Net has low model complexity, in which the parameters, computation cost, and inference speed are comparable to the SENet. We evaluate BA-Net on three computer vision tasks: image classification, object detection, and instance segmentation. The BA-Net has achieved higher performance compared to other channel attention methods. 

In general, the main contributions of this manuscript can be summarized as follows:
\begin{itemize}
\item We demonstrate the limitation of traditional channel attention mechanisms with theoretical analysis and propose the Bridge Attention Net (BA-Net) as a new solution, in which richer information from previous convolution layers is bridged and well utilized. 
\item With extensive evaluation, we conclude that the closer convolution outputs are more effective. We bridge all the convolution outputs inside each block, thus get the best BA-Net structure.
\item Experimental evaluation shows that the proposed BA-Net achieves the higher performance on various computer vision tasks among attention methods. Based on the ResNet-50 backbone, it significantly outperforms SENet50 by 0.71\% in terms of TOP-1 accuracy with affordable model complexity and fast speed.

\end{itemize}

\section{Related Work}
We mainly revisit attention mechanisms and cross-layer interaction applied in Convolutional Neural Networks(CNNs) in the existing literature.

\textbf{Attention mechanisms.} The attention mechanism is capable of enhance the assignment of the most informative feature representations while suppressing the less useful ones, thus allowing the model to focus on the important regions in the context adaptively. The pioneering SENet\cite{hu2018squeeze} is the cornerstone of the attention mechanism research field. The method extracted channel-wise features by simple global average pooling and full connection layers and this significantly improved the performance of many CNNs with little parameters and computing cost added. The SKNet \cite{li2019selective} enhances the expressiveness of the model by passing the feature map through two convolution layers of different kernel sizes, followed by the extraction of channel attention. While extracting channel attention, BAM \cite{park2018bam} and CBAM \cite{woo2018cbam} utilize the spatial information and generate spatial attention using convolution. DA-Net Attention \cite{fu2019dual} concentrates on relevance of local and global features, and combines the two features by summing the attention modules of two branches. ResNeSt \cite{zhang2020resnest} adopts a similar split-attention block, which enables the fusion of attention between different groups of the input feature maps. GSoPNet proposes the global second-order pooling to introduce higher-order representation for the improving non-linear capability of CNNs. GSoPNet \cite{gao2019global} obtains attention by fully using the second-order statistics of the holistic image. Fca-Net \cite{qin2020fcanet} revisits channel attention using frequency analysis and generalizes the pre-processing of channel attention mechanism in the frequency domain. Some methods explore lightweight strategies to reduce the parameters and computing cost of the model with attention. ECA-Net Attention \cite{wang2020eca} proposes local cross-channel interaction and generates attention by 1D convolution. SA-Net is also a lightweight attention structure inspired by channel shuffling. Half of the features are used to generate spatial attention in SA-Net, and the other half are used to generate channel attention.  At the end of the block, features are shuffled along channel. The above mechanisms provide many novel ways to generate channel or spatial attention. However one thing in common among them is that they only focus on the features of the layer adjacent to attention layer The features in previous layers are ignored. In this paper we will explore the effect of features in previous layers on the attention mechanism.

\textbf{Cross-layer integration.} It is a common strategy to improve network representation by skip connection, which can solve the problem of gradient dispersion and the disappearance of deep networks to some extent. This strategy can train deep networks more adequately, making deeper network structures feasible, and has been widely used in the design of models. The ResNet \cite{he2016deep}network first proposed a residual module, which facilitates the fusion of information between different layers. DenseNet \cite{huang2017densely} also uses a similar structure but differs from ResNet \cite{he2016deep}in the form of concatenating for feature stitching. In U-Net \cite{ronneberger2015u}, which is commonly used in the field of medical segmentation, the decoder-encoder module is connected through skip connection to make feature extraction reach a higher accuracy. 
 
Actually, cross-layer integration has been used to improve the performance of attention mechanisms. Duo L. et al. \cite{li2020deep} proposes the DREAL method to optimize parameters of arbitrary attention modules, in which LSTM\cite{hochreiter1997long} is used to integrate previous attention weights and deep reinforce learning is used to update parameters of LSTM and attention layers. DIANet \cite{huang2020dianet} also utilizes LSTM module to integrate previous attention weights and directly outputs attention weights in current block by LSTM. DIANet visualizes the effect of previous features acting on the current attention layer and shows the effect on stabilizing Training. Yu. W. et al.\cite{wang2021evolving} proposes the evolving attention to improve the performance of transformers, named as EA-AANet. Attention maps in a preceding block are integrated with ones in current layer by residual connection and 2D convolution. Compared to these works, the proposed BA-Net in this paper has the similar motivation, but this approach integrates the features in previous layers of the current block. The higher performance of our models displayed in Table.\ref{ImageNet} demonstrates that feature integration of our method is more effective.

\section{Approach}
In this section, we first review traditional channel attention mechanisms(i.e., SENet\cite{hu2018squeeze}). We give out the common form of the mechanisms and demonstrate the limitation of the mechanisms through theoretical analysis. It inspires us to come up with the Bridge Attention mechanism, and we will concretely introduce the implementation of the proposed module. 

\subsection{Traditional channel attention mechanisms}
\subsubsection{Common form.}
We demonstrate the common form of attention mechanisms by SENet. Let the output of the SE block be $X \in \mathbb{R}^{C\times H\times W}$, where $C$, $H$ and $W$ are channel, height and width dimension of the output. Accordingly, the generated attention weights can be computed as:
\begin{equation}
\omega = \sigma (\mathcal{F_C}(gap(X))) \label{SE}
\end{equation}
where $gap(X)=\frac{1}{HW}\sum_{i=1,j=1}^{H,W}X_{i,j}$ is channel-wise global average pooling, $\sigma(\cdot)$ represents Sigmoid function. $\mathcal{F_C}(\cdot)$ represents two stacked Full Connection(FC) layers, which can be expressed as followed:
\begin{equation}
\mathcal{F_C}(y) = (\bm{W_2}) ReLU (\bm{W_1} y) \label{FC}
\end{equation}
In Eqn.\ref{FC}, ReLU \cite{nair2010rectified} represents an activation function. $\bm{W_1}$ and $\bm{W_2}$ are matrix used to form the attention weights. The two matrix respectively have size of $C \times (\frac{C}{r})$ and $(\frac{C}{r}) \times C$, in which the reduction factor $r$ is used to avoid heavy computation and high complexity of the attention layer.

We consider that attention mechanism can be divide into two parts, \textbf{Integration} and \textbf{Generation}. In SENet, the output $X$ is first squeezed by average pooling $gap(\cdot)$ and fully integrated among channels by matrix $\bm{W_1}$, considered as \textbf{Integration} $\mathcal{I}(\cdot)$. And then the features are sequentially fed into $RELU$, matrix $\bm{W_2}$, $\sigma(\cdot)$, to get the final attention weights, considered as \textbf{Generation} $\mathcal{G}(\cdot)$. Thus the form of attention mechanism can be expressed as :
\begin{align}
\mathcal{I}(\cdot) = \bm{W_1}(gap(\cdot)),\  &\mathcal{G}(\cdot) = \sigma(\bm{W_2}(ReLU(\cdot))) 
&\\ \Longrightarrow \omega = \mathcal{G}(\mathcal{I}(X)) \label{common form}
\end{align}
In our method, richer features of previous layers are bridged and integrated in $\mathcal{I}(\cdot)$ beside features of adjacent layer, thus the generated attention weights should be more adapted to the output $X$.

\subsubsection{Limitation.}
Fig.\ref{Most Attention Module} shows the block architecture of most existing attention methods, which includes the convolution part and an attention layer. Let $\mathcal{F}(\cdot)$ and $att(\cdot)$ represent the convolution part and the attention layer respectively, thus the whole process can be expressed as:
\begin{equation}
    \mathcal{F}_{att}(\cdot) = \mathcal{F}(\cdot) \odot att(\mathcal{F}(\cdot)) \label{Fatt}
\end{equation}
$\odot$ represents the element-wise multiplication.

Generally, the convolution part consists of several stacked convolution layers. We assume that the total number of convolution layers is $n$, thus:
\begin{equation}
    \mathcal{F}(\cdot) = F_n(F_{n-1}(\cdots F_2(F_1))) \label{Fstack}
\end{equation}
$F_i(\cdot)$ represents the certain convolution layer, where $1\leq i \leq n$. 

Considering the distance, we assume that the outputs of $\mathcal{F}(\cdot)$ are more implicitly correlated with the previous $q$ layers,  thus Eqn.\ref{Fstack} can be approximately equal to:
\begin{align}
    \mathcal{F}(\cdot) \approx F_n(F_{n-1}(\cdots F_{n-(q-1)}(F_{n-q})))\\
    \Rightarrow \mathcal{F}(F_n, F_{n-1}, \dots, F_{n-(q-1)}, F_{n-q})
\end{align}

In most existing attention methods, the attention layer only takes the outputs of the adjacent layer $F_n(\cdot)$. Some methods even have complicated calculations in $att(\cdot)$ for richer information, which weakens the correlation with the previous $q$ convolution layers: 
\begin{equation}
    att(\cdot) = att(F_n(F_{n-1}(\cdots F_{n-(q-1)}(F_{n-q})))) \approx att(F_n) \label{weaken}
\end{equation}
According to Eqn.\ref{weaken}, the generated attention weights lack correlation with previous layers, resulting in insufficient adaptive to outputs of $\mathcal{F}(\cdot)$.

In fact, \cite{huang2020dianet} \cite{wang2021evolving} have noticed the above issue, but only the attention weights of previous blocks are fed into current attention layer. In our method, features of more previous layers are bridged to current attention layer, while the distance is shorter,  thus providing more effective information.

\subsection{Bridge Attention mechanism}
In this part, we concretely introduce how previous features are integrated in our method and give out the implementation of Bridge Attention module. Let the output of $F_i(\cdot)$ inside the block be $X_i \in \mathbb{R}^{C_i\times H\times W}$. The outputs are first global average pooled ($gap$) to the size of $C_i\times 1\times 1$, and then fed into respective matrices of size $C_i \times (\frac{C_n}{r})$ to get the squeezed features. The size $\frac{C_n}{r}\times 1\times 1$ is the same as the squeezed feature from the output of $F_n(\cdot)$, which is followed by $att(\cdot)$. Thus the squeezed features from different layers are directly added and the final integrated feature is obtained. 

We notice that distributions among the squeezed features can be in massive difference due to $gap$ and their squeezing matrix, so we apply Batch Normalization for the features to make them in similar distributions, thus the integration can be more effective. In addition, Batch Normalization improves the nonlinear representation of the features, which benefits network parameters updating. In a whole, the integration part can be expressed as:
\begin{align}
    &S_i = BN_i(\bm{W_i}(gap(F_i))\\
    &\mathcal{I}_{BA}(\cdot) = \sum_{n-q}^{n}{S_i}
    \label{Bridge intrinsic block}
\end{align}
$S_i$ represents the squeezed feature from $F_i(\cdot)$. Then we input the integrated feature into generation part and get the final attention weights:
\begin{align}
    \omega = \mathcal{G}(\mathcal{I}_{BA}(\cdot))
\end{align}

\begin{figure*} 
\centering
\includegraphics[height=4.9cm, width=17cm]{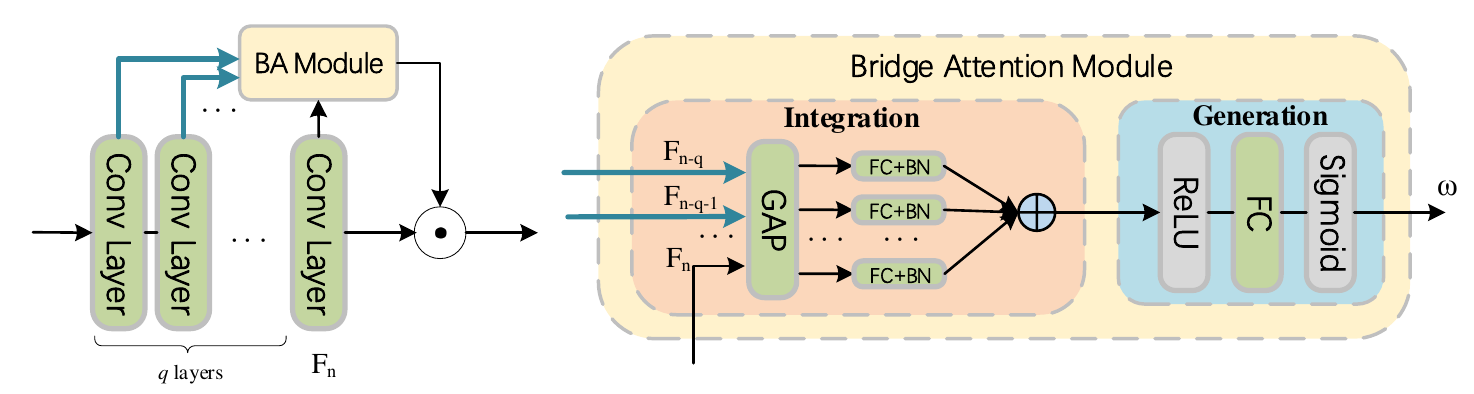}\\ 
\caption{The overview of Bridge Attention module. Blue arrows indicate that features from previous $q$ layers inside the block are bridged to the attention layer. \textbf{FC} indicates the matrix to squeeze features and \textbf{BN} is BatchNorm Layer. }\label{Bridge Attention Module} 
\end{figure*}

\subsection{Implementation of BA module}
Similar methods only use previous attention weights as addition features to the attention layer. However, there are various previous features can be bridge to the attention layer. So to investigate the effectiveness of different previous features. we have evaluation on bridging previous attention weights, preivous convolution outputs at different positions, respectively. We take the experiment base on ResNet-50 and the result is shown in \ref{diff_bridge}. 

The row$_3$ achieves higher accuracy than the row$_2$, although they are at the same position. Because the attention weights are used to rescale the feature maps, compared with the convolution outputs, they contribute less helpful information to the attention layer. Besides, bridging the proximate features can achieve better performance. Due to the heavy feature compression in the attention layer, more proximate features can straightforwardly supplement more relevant information. 

As a conclusion, bridging the closer convolution outputs can achieve better performance. So we merely consider the features within the block, avoiding a significant increase in configuration complexity. Generally, the blocks of existing CNNs contain no more than three convolution layers, so we bridge all convolution outputs before the attention layer. The values of $n$ and $q$ can be determined by the position of the attention layer in the block. For the block of ResNet, the attention layer is usually placed after the third convolution layer, thus $n=3, q=2$. The block of MoblileNetv3 or EffcientNet already contains an attention layer, which follows the second convolution layer, thus $n=2, q=1$. In addition, we also set $n=2, q=1$ for the block with two convolution layers, such as the plain backbone of ResNet. The application of our method to convolution block is shown in Fig.\ref{Bridge Attention Module}.

\begin{table}[]
    \centering
    \caption{The comparison when bridging different features. $att$: the attention weights, $conv$: the convolution outputs, $prev$: the previous block, $curr$: the current block, $conv_i$: the i-th convolution layer, $end$: the end of the block.}
    \begin{tabular}{l | c | c | c  }
    \hline
    Backbone &Type &Position &TOP-1(\%) \\
    \hline
    SE ResNet-50 &$None$ &--- &78.14\\
    \hline
    \makecell[l]{BA ResNet-50}
    &\makecell[c]{$att$\\ $conv$\\ $conv$\\ $conv$\\ $conv$\\ $conv$}
    &\makecell[c]{$prev, conv_3$\\ $prev, conv_3$\\ $prev, end$\\ $curr, conv_1$\\ $curr, conv_2$\\ $curr, conv_{1\&2}$}
    &\makecell[c]{78.41\\ 78.49\\ 78.54\\ 78.78\\ 78.77\\ \textbf{78.85}}\\
    %&\makecell[c]{94.20\\ 94.24\\ 94.17\\ 94.34\\ 94.31\\ 94.28}
    \hline
    \end{tabular}
    \label{diff_bridge}
\end{table} 

\section{Experiments}
In this section, we evaluate our method on three computer vision tasks, including image classification, object detection, and instance segmentation. We first demonstrate the implementation details of the experiments. Then, we give out the performance comparison of our method with other attention methods.

\subsection{Implementation details}
For image classification, we evaluate the performance on ImageNet-1K\cite{russakovsky2015imagenet} dataset, where we apply our method on various backbone architectures, including ResNet \cite{he2016deep}, MobileNet-v3, EfficientNet and ResNeXt. We take the same strategy of the data augmentation, and hyperparameter settings in \cite{he2016deep} and \cite{he2019bag}. The training images are cropped randomly to 224×224 with random horizontal flipping, while the testing images are resized to 256×256 and cropped from center to 224×224. We use an SGD optimizer with a momentum of 0.9 and a weight decay of 1e-4. In the training phase, the initial learning rate is set to 0.1 for a batch size of 256. All models are trained within 100 epochs with
cosine learning rate decay following FcaNet\cite{qin2020fcanet}.

For object detection and instance segmentation, we evaluate our method on the MS COCO2017 dataset \cite{lin2014microsoft}. Faster R-CNN\cite{ren2016faster} and  Mask R-CNN\cite{he2017mask} are used as detectors while BA-Net-50 \& 101 pretained on ImageNet-1K are used as backbone. We used MMDetection toolkit\cite{chen2019mmdetection} to implement all detectors and follow the default settings. The shorter side of input images are resized to 800. All models are optimized using SGD with weight decay of 1e-4, momentum of 0.9, and batch size is set to 8. The total number of training epochs is 12, and the initial learning rate is 0.01, decreased by a factor of 10 at the 8th and 11th epoch, respectively.

We construct all models based on the PyTorch framework and experiment on four Nvidia RTX 3090Ti GPUs.

\subsection{Image Classification on ImageNet-1K}
\subsubsection{Performance comparison with other methods.}Firstly, we evaluate our method under the backbones of ResNet-50\&101, which are the most common backbones used to apply attention mechanisms. Besides traditional channel mechanisms like SENet, ECA-Net, FcaNet,  we also compare the performance with the methods using cross-layer integration, like DREAL, DIANet and EA-AANet. We give out the metrics from their origin papers. In addition, we noticed that different training settings are used in different mechanisms in their origin papers, so we retrained part of attention models that is reproducible, following the setting of FcaNet\cite{qin2020fcanet}. Observed in Table.\ref{ImageNet}, BA-Net has higher performance than other attention mechanisms in $org.$ metrics, specifically BA-Net significantly outperforms SENet by 2.14\% and 1.41\% in $org.$ TOP-1 under the backbones of ResNet-50 and ResNet-101, respectively. Under the same training setting, our method also performs better than SENet, ECA-Net and FcaNet, specifically BA-Net outperforms SENet by 0.71\% and 0.62\% in $self.$ TOP-1 under the two backbones, respectively.
\begin{table*}
	\centering
	\caption{Performance comparisons of different attention methods on ImageNet-1K in terms of network parameters(Param.), floating point operations per second (FLOPs), and Top-1/Top-5 accuracy. The term $self.$ means that the metrics came from the experiments retrained by ourselves while $org.$ means that the metrics came from the origin paper. Our method and the best records are marked in \textbf{bold}.}
	\begin{tabular}{l | c | c | c | c | c }
	\hline
	\rule{0pt}{10pt}
    Attention Method  &Backbone	&Param. &FLOPs &\makecell[c]{TOP-1(\%)\\ \hline $self. \ |\ \   org.$}  &\makecell[c]{TOP-5(\%)\\ \hline $self. \ |\ \   org.$}\\
	\hline
	\rule{0pt}{20pt}
 	\makecell[l]{ResNet\\ SENet\\ ECA-Net\\ FcaNet\\ SENet+DREAL\\ DIANet\\ EA-AANet\\ \textbf{BA-Net(ours)}} 	%Method
	&\makecell[c]{ResNet-50}
	&\makecell[c]{25.56M\\ 28.07M\\ 25.56M\\ 28.07M\\ 28.12M\\ 28.38M\\ 25.80M\\ 28.71M}
	&\makecell[c]{4.12G\\ 4.13G\\ 4.13G\\ 4.13G\\ 4.13G\\ \textbf{---}\\ 4.35G\\ 4.13G}
	&\makecell[c]{
	 \ \ \textbf{---} \ $|$\ 75.20\\ 
	 78.14 $|$\ 76.71\\ 
	 77.98 $|$\ 77.43\\ 
	 78.57 $|$\ 78.52\\ 
	 \ \ \textbf{---} \ $|$\ 77.85\\ 
	 \ \ \textbf{---} \ $|$\ 77.24\\ 
	 \ \ \textbf{---} \ $|$\ 78.22\\ 
	\textbf{78.85}}	
	&\makecell[c]{
	\ \ \textbf{---} \ $|$\ 92.52\\ 
	 94.05 $|$\ 93.38\\ 
	 93.94 $|$\ 93.65\\ 
	 94.16 $|$\ 94.14\\ 
	 \ \ \textbf{---} \ $|$\ 94.05\\ 
	 \textbf{---} \\
	 \ \ \textbf{---} \ $|$\ 94.21\\ 
	\textbf{94.28}}\\
	\hline
	\rule{0pt}{20pt}
	\makecell[l]{ResNet\\ SENet\\ ECA-Net\\ FcaNet\\ SENet+DREAL\\ EA-AANet\\ \textbf{BA-Net(ours)}}
	&\makecell[c]{ResNet-101}
	&\makecell[c]{44.55M\\ 49.29M\\ 44.55M\\ 49.29M\\ 49.36M\\ 45.40M\\ 50.49M}	
	&\makecell[c]{7.85G\\ 7.86G\\ 7.87G\\ 7.86G\\ 7.87G\\  8.60G\\ 7.87G}
	&\makecell[c]{
     \ \ \textbf{---} \ $|$\ 76.83\\ 
	 79.41 $|$\ 77.62\\ 
	 79.23 $|$\ 78.65\\ 
	 79.63 $|$\ 79.64\\ 
	 \ \ \textbf{---} \ $|$\ 79.27\\ 
	 \ \ \textbf{---} \ $|$\ 79.29\\ 
	\textbf{80.03}}
	&\makecell[c]{
	\ \ \textbf{---} \ $|$\ 93.48\\ 
	 94.62 $|$\ 93.93\\ 
	 94.45 $|$\ 94.34\\ 
	 94.66 $|$\ 94.63\\ 
	 \ \ \textbf{---} \ $|$\ 94.59\\ 
	 \ \ \textbf{---} \ $|$\ 94.81\\ 
	\textbf{94.83}}\\
	\hline
	\end{tabular}
	\label{ImageNet}
\end{table*}

\subsubsection{Computing cost. } Obeserved in Table.\ref{ImageNet}, parameters of BA-Net are slightly larger than parameters fo SENet since the features of previous layers are bridged to the attention layer, while FLOPs of them are almost the same. To further illustrate computation cost of BA-Net, we make additional comparisons on graphics memory usage and speed when training and testing. In Table.\ref{SENet}, memory usages when training and testing are slightly increased while training speed and testing speed are slightly decreased. With comparable computing cost, BA-Net outperforms SENet by 0.71\% and 0.62\% under the two backbones.
\begin{table*}
	\centering
	\caption{Computing cost comparisons of BA-Net and SENet in temrs of memory usage and speed(frame per second, FPS) when train and test. M. represents memory usage and S. represents speed.}
	\begin{tabular}{l | c | c | c | c | c | c}
	\hline
	\rule{0pt}{10pt}
	Method &Backbone	&Train M.   &Train S.   &Test M.    &Test S.    &TOP-1(\%)\\
	\hline
	\rule{0pt}{16pt}
	\makecell[l]{SENet\\ \textbf{BA-Net(Ours)}}
	&\makecell[c]{ResNet-50}
	&\makecell[c]{34.74G\\ 34.85G}
	&\makecell[c]{656 FPS\\ 612 FPS}
	&\makecell[c]{7.49G\\ 7.67G}
	&\makecell[c]{1315 FPS\\ 1280 FPS}
	&\makecell[c]{78.14\\ \textbf{78.85}}\\
	\hline
	\rule{0pt}{16pt}
	\makecell[l]{SENet\\ \textbf{BA-Net(Ours)}}
	&\makecell[c]{ResNet-101}
	&\makecell[c]{46.86G\\ 47.54G}
	&\makecell[c]{397 FPS\\ 362 FPS}
	&\makecell[c]{8.96G\\ 9.26G}
	&\makecell[c]{845 FPS\\ 792 FPS}
	&\makecell[c]{79.41\\ \textbf{80.03}}\\
	\hline
	\end{tabular}
 	\label{SENet}
\end{table*}

\subsubsection{Application on other backbones. }To further verify the capability of Bridge Attention on other backbone architectures, we apply our method on ResNext, EfficientNet, and MobileNetv3. We retrained all the original backbones while the types with Bridge Attention added, comparisons are shown in Table.\ref{application}. For ResNeXt-50, BA improves the model by 1.19\% on TOP-1 and 0.51\% on TOP-5. EfficientNet-b0 with BA significantly outperforms the origin model by 1.59\% and 0.76\% on TOP-1 and TOP-5, respectively. However, Bridge Attention seems to have little effect on light-weight backbone like MoblileNetv3, with only a 0.21\% improvement on TOP-1 of the large type. So we consider that BA can help improve performance more significantly on heavy backbone architectures. 
\begin{table*}
    \centering
    \caption{Performance comparisons of BA-Net application on different backbone architectures.}
    \begin{tabular}{l | c | c | c | c }
    \hline
    \rule{0pt}{10pt}
    BackBone   &Type   &Param. &TOP-1(\%)  &TOP-5(\%)\\
    \hline
    \rule{0pt}{10pt}
    \makecell[l]{ResNeXt-50}
    &\makecell[c]{Origin\\ \textbf{+BA}}
    &\makecell[c]{25.05M\\ 28.8M}
    &\makecell[c]{78.77\\ 79.58}
    &\makecell[c]{94.18\\ 94.69}\\
    \hline
    \rule{0pt}{10pt}
    \makecell[l]{EfficientNet-b0}
    &\makecell[c]{Origin\\ \textbf{+BA}}
    &\makecell[c]{15.1M\\ 16.2M}
    &\makecell[c]{70.11\\ 71.70}
    &\makecell[c]{89.45\\ 90.21}\\
    \hline
    \rule{0pt}{10pt}
    \makecell[l]{MobileNetv3-small}
    &\makecell[c]{Origin\\ \textbf{+BA}}
    &\makecell[c]{2.1M\\ 2.3M}
    &\makecell[c]{65.87\\ 65.85}
    &\makecell[c]{86.26\\ 86.48}\\
    \hline
    \rule{0pt}{10pt}
    \makecell[l]{MobileNetv3-large}
    &\makecell[c]{Origin\\ \textbf{+BA}}
    &\makecell[c]{3.5M\\ 4.1M}
    &\makecell[c]{73.29\\ 73.50}
    &\makecell[c]{91.18\\ 91.22}\\
    \hline
    \end{tabular}
    \label{application}
\end{table*}

\subsection{Object detection on COCO2017}
In this subsection, we evaluate our BA-Net on object detection task using Fast R-CNN and Mask R-CNN. We mainly compare BA-Net with ResNet, SENet, ECA-Net and FcaNet. We transferred our BA-Net models on the COCO2017 training set and gave out the metrics tested on the validation set.  As shown in Table.\ref{COCO_detection}, most metrics of BA-Net achieve the highest performance. For Fast R-CNN, our BA-Net outperforms SENet by 1.8\% and 2.1\% in terms of $mAP$ with the backbones ResNet-50 and ResNet-101, respectively. For Mask R-CNN, our BA-Net outperforms SENet by 2.1\% in terms of $mAP$ with ResNet-50.
\begin{table*}[]
    \centering
    \caption{Performance comparisons of different attention methods on obeject detection task. Average Precision($AP$) is the main comparison metric.}
    \begin{tabular}{l | c | c | c | c | c | c | c | c | c}
    \hline
    Backbone &Detector &Param. &FLOPs &$mAP$ &$AP_{50}$ &$AP_{75}$ &$AP_S$ &$AP_M$ &$AP_L$\\
    \hline
    \makecell[l]{ResNet-50\\ SENet\\ ECA-Net\\ FcaNet \\ \textbf{BA-Net}\\ 
    \hline 
    ResNet-101\\ SENet\\ ECA-Net\\ FcaNet \\ \textbf{BA-Net}\\}
    &\makecell[c]{Faster-RCNN}
    &\makecell[c]{41.53M\\ 44.02M\\ 41.53M\\ 44.02M\\ 44.66M\\ 
    \hline 
    60.52M\\ 65.24M\\ 60.52M\\ 65.24M\\ 66.44M}
    &\makecell[c]{215.51G\\ 215.63G\\ 215.63G\\ 215.63G\\ 215.68G\\ 
    \hline 
    295.39G\\ 295.58G\\ 295.58G\\ 295.58G\\ 295.70G}
    &\makecell[c]{36.4\\ 37.7\\ 38.0\\ 39.0\\ \textbf{39.5}\\
    \hline
    38.7\\ 39.6\\ 40.3\\ 41.2\\ \textbf{41.7}}
    &\makecell[c]{58.2\\ 60.1\\ 60.6\\ 61.1\\ \textbf{61.3}\\
    \hline
    60.6\\ 62.0\\ 62.9\\ 63.3\\ \textbf{63.4} }
    &\makecell[c]{39.2\\ 40.9\\ 40.9\\ 42.3\\ \textbf{43.0}\\
    \hline
    41.9\\ 43.1\\ 44.0\\ 44.6\\ \textbf{45.1}}
    &\makecell[c]{21.8\\ 22.9\\ 23.4\\ 23.7\\ \textbf{24.5}\\
    \hline
    22.7\\ 23.7\\ 24.5\\ 23.8\\ \textbf{24.9}}
    &\makecell[c]{40.0\\ 41.9\\ 42.1\\ 42.8\\ \textbf{43.2}\\
    \hline
    43.2\\ 44.0\\ 44.7\\ 45.2\\ \textbf{45.8}}
    &\makecell[c]{46.2\\ 48.2\\ 48.0\\ 49.6\\ \textbf{50.6}\\
    \hline
    50.4\\ 51.4\\ 51.3\\ 53.1\\ \textbf{54.0}}\\
    \hline
    \makecell[l]{ResNet-50\\ SENet\\ ECA-Net\\ FcaNet \\ \textbf{BA-Net}}
    &\makecell[c]{Mask-RCNN}
    &\makecell[c]{44.17M\\ 46.66M\\ 44.17M\\ 46.66M\\ 47.30M}
    &\makecell[c]{261.81G\\ 261.93G\\ 261.93G\\ 261.93G\\ 261.98G}
    &\makecell[c]{37.2\\ 38.4\\ 39.0\\ 40.3\\ \textbf{40.5}}
    &\makecell[c]{58.9\\ 60.9\\ 61.3\\ \textbf{62.0}\\ 61.7}
    &\makecell[c]{40.3\\ 42.1\\ 42.1\\ 44.1\\ \textbf{44.2}}
    &\makecell[c]{22.2\\ 23.4\\ 24.2\\ \textbf{25.2}\\ 24.5}
    &\makecell[c]{40.7\\ 42.7\\ 42.8\\ 43.9\\ \textbf{44.3}}
    &\makecell[c]{48.0\\ 50.0\\ 49.9\\ 52.0\\ \textbf{52.1}}\\
    \hline
    \end{tabular}
    \label{COCO_detection}
\end{table*}
\subsection{Instance segmentation on COCO2017}
For instance segmentation task, we take Mask R-CNN as the detector for evaluation and the result is shown in Table.\ref{COCO_instance}. $mAP$ of BA-Net achieved 36.6\% and 37.8\% under the two backbones, which performs better than other attention methods. Compared with SENet, our BA-Net notably outperformed by 1.2\% and 1.3\% in terms of $mAP$, respectively. Besides image classification, BA-Net also performs well on object detection and instance segmentation tasks, which verifies that our BA-Net has good generalization ability for various tasks.
\begin{table*}[]
    \centering
    \caption{Performance comparisons of different attention methods on instance segmentation task.}
    \begin{tabular}{l | c | c | c | c | c | c }
    \hline
    Backbone &$mAP$ &$AP_{50}$ &$AP_{75}$ &$AP_S$ &$AP_M$ &$AP_L$\\
    \hline
    \makecell[l]{ResNet-50\\ SENet\\ ECA-Net\\ FcaNet \\ \textbf{BA-Net}\\ 
    \hline
    ResNet-101\\ SENet\\ ECA-Net\\ \textbf{BA-Net}\\}
    &\makecell[c]{34.2\\ 35.4\\ 35.6\\ 36.2\\ \textbf{36.6}\\
    \hline
    35.9\\ 36.8\\ 37.4\\ \textbf{38.1}}
    &\makecell[c]{55.9\\ 57.4\\ 58.1\\ 58.6\\ \textbf{58.7}\\ 
    \hline
    57.7\\ 59.3\\ 59.9\\ \textbf{60.6}}
    &\makecell[c]{36.2\\ 37.8\\ 37.7\\ 38.1\\ \textbf{38.6}\\
    \hline
    38.4\\ 39.2\\ 39.8\\ \textbf{40.4}}
    &\makecell[c]{\textbf{18.2}\\ 17.1\\ 17.6\\ ---\\ \textbf{18.2}\\
    \hline
    16.8\\ 17.2\\ 18.1\\ \textbf{18.7}}
    &\makecell[c]{37.5\\ 38.6\\ 39.0\\ ---\\ \textbf{39.6}\\
    \hline
    39.7\\ 40.3\\ 41.1\\ \textbf{41.5}}
    &\makecell[c]{46.3\\ 51.8\\ 51.8\\ ---\\ \textbf{52.3}\\
    \hline
    49.7\\ 53.6\\ 54.1\\ \textbf{54.8}}\\
    \hline
    \end{tabular}
    \label{COCO_instance}
\end{table*}
\section{Analysis}
\subsection{Effectiveness of Bridge Attention}
\begin{figure*} 
\centering
\includegraphics[height=4.6cm, width=16cm]{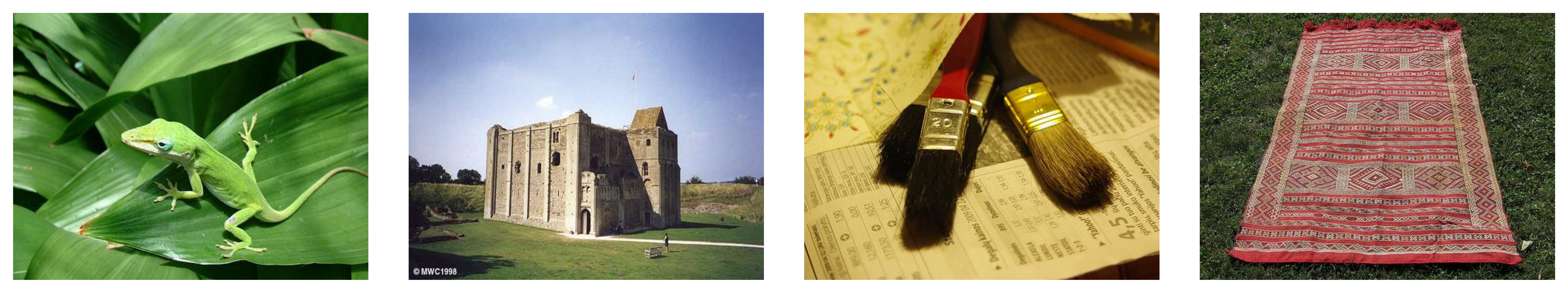}\\ 
\caption{Example images of four classes from ImageNet. The images from left to right are American chameleon, castle, paintbrush and prayer mat, respectively.}\label{Examples of four classes.} 
\end{figure*}

\begin{figure*} 
\centering
\includegraphics[height=8.5cm, width=17cm]{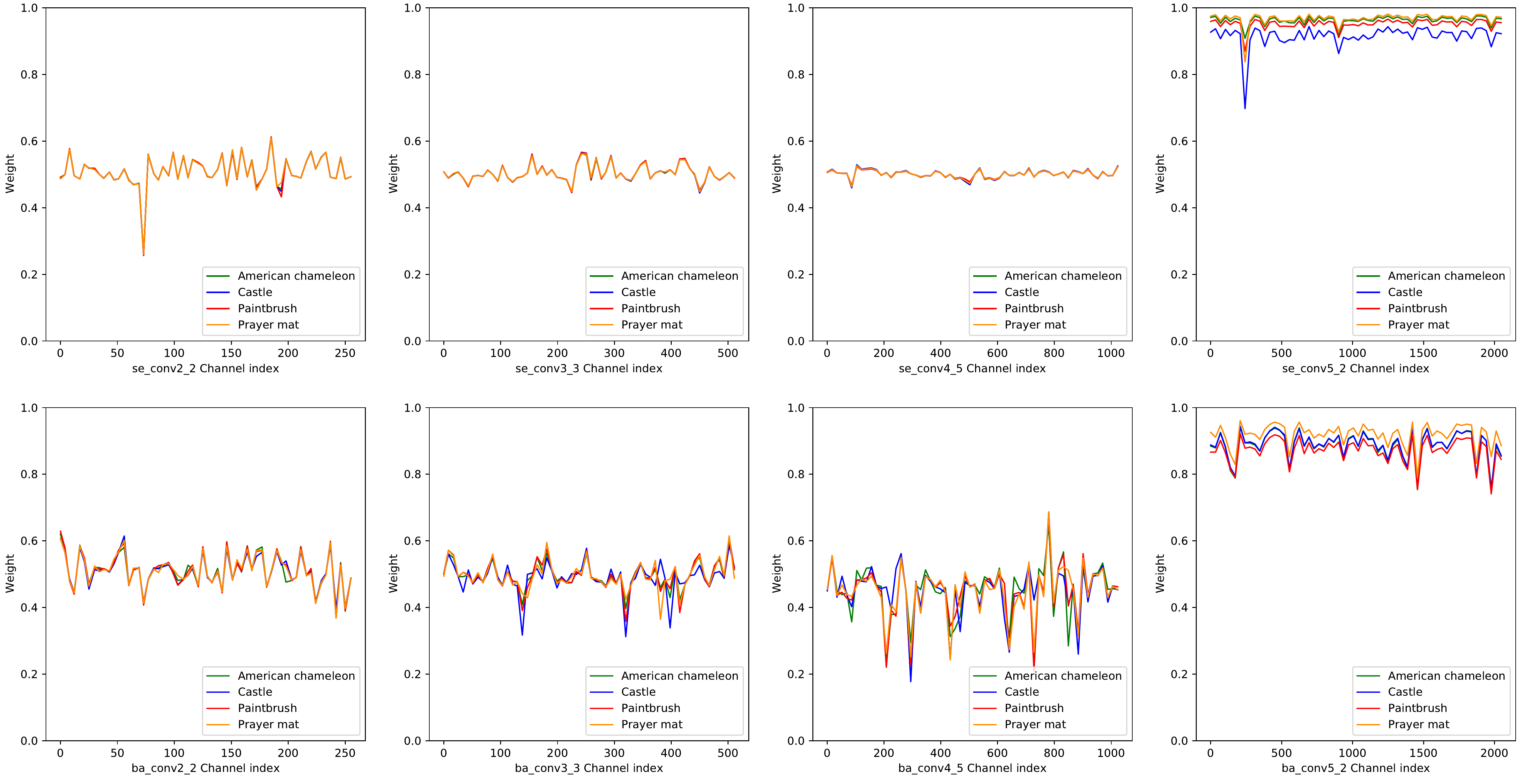}\\ 
\caption{Visualization of chaneel attention weights of Block-$i_j$, where $i$ indicate the $i$-th stage and $j$ is $j$-th block in $i$-th stage. The weights learned by SENet50 blocks and BA-Net50 are illustrated in top and bottom row, respectively.}\label{Attention weights visualization} 
\end{figure*}

To further analyze how Bridge Attention affects the feature map, we visualize the attention weights distribution of BA-Net and compare it with SENet. Concretely, we randomly sample four classes from ImageNet, which are American chameleon, castle, paintbrush and prayer mat, respectively. All images of each class are collected from the validation set of ImageNet, and some example images are shown in Figure \ref{Examples of four classes.}. We put the images of the same class into the pretrained BA-Net-50 and SENet-50, then compute the channel attention weights of convolution blocks on average. Figure \ref{Attention weights visualization} visualizes the attention weights of four blocks, and each is the last block of four stages. The attention weights of SE blocks are illustrated in the top row, while BA blocks' are illustrated in the bottom row. % conv $i\_j$, which indicates $j$-th convolution block in $i$-th stage
In the first stage, the weights distributions of both models are similar, showing that enhancement from Bridge Attention is not significant for coarse feature extraction. But in the later stages for detailed feature extraction, the variance of the weight distribution in BA-Net increases significantly, indicating that the weights become more diverse , especially in the third stage. It demonstrates that BA-Net can effectively capture the more important features while filtering out the less important ones, thus enhancing the  representation of the feature maps. In addition, the SENet's weights curves of different classes in the first three stages almost overlap, while the BA-Net's curves of different classes are clearly distinguishable on some channels. This indicates that BA-Net is able to distinguish detailed features of different classes sharply. In general, the bridged features of previous convolution layers effectively enhance the representation ability of the output feature maps.
\subsection{Importance of the integrated features}
In our method, the features of different convolution layers in the block are integrated into the attention layer, so we want to explore the relationship between the integrated features and the attention weights and how the features contribute to the attention weights. We consider using the random forest model to reveal the relationship and take the Gini importance\cite{gregorutti2017correlation} from the model as the measurement of feature importance. 
\begin{figure*} 
\centering
\includegraphics[height=4.36cm, width=16cm]{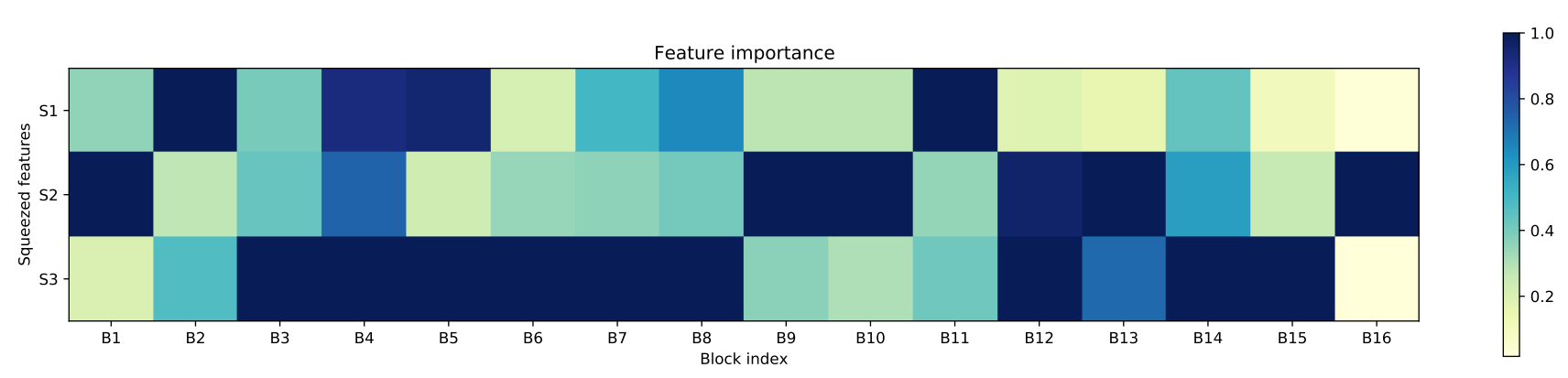}\\ 
\caption{Visualization of Gini importance of the integrated features. B$i$ indicates the $i$-th block in BA-Net50 model and S$i$ is $i$-th squeezed feature in a certain block. The deeper the color of the square, the more important it is.}\label{feature importance} 
\end{figure*}

There are 16 blocks in the BA-Net-50, where each block contains three convolution layers, and the attention layer follows the third layer. We also use the validation set of ImageNet to inference in the model, and then we get the squeezed features $S^i$ and attention weights $\omega^i$ in each block. We fit a random forest model for each block using its $S^i$ and $\omega^i$, and then visualize the feature importance as shown in Fig.\ref{feature importance}. We notice that not all $S_3$ contribute the most to the attention weights, which is adjacent to the attention layer. For example, the contribution of $S_2$ is comparable to $S_3$ in B12, or even the integrated features from previous convolution layers are more than important $S_3$, such as B11, B16. The results demonstrate that the features from previous convolution layers also effectively contribute to the attention weight even play a dominant role among the integrated features in a certain block.\\

\section{Conclusion}
Traditional channel attention methods use only the output of the adjacent convolution layer. Faced with this limitation, this paper proposes a novel idea named Bridge Attention to enrich the information for better channel weight estimation. We design Bridge Attention Module with simple strategies to make this idea can be easily applied in various AI tasks.  Experimental evaluation shows that significant performance improvement over the existing state-of-the-art methods can be achieved by integrating the features from the previous convolution layers inside the block. In the future work, we will consider extending the Bridge Attention exploring feature integration from the previous block, thus further improving the neural network's performance.

\bibstyle{aaai22.bst}
\bibliography{BA-Net_arxiv.bib}

\begin{thebibliography}{30}
\providecommand{\natexlab}[1]{#1}

\bibitem[{Bello et~al.(2019)Bello, Zoph, Vaswani, Shlens, and
  Le}]{bello2019attention}
Bello, I.; Zoph, B.; Vaswani, A.; Shlens, J.; and Le, Q.~V. 2019.
\newblock Attention augmented convolutional networks.
\newblock In \emph{Proceedings of the IEEE/CVF international conference on
  computer vision}, 3286--3295.

\bibitem[{Chen et~al.(2019)Chen, Wang, Pang, Cao, Xiong, Li, Sun, Feng, Liu, Xu
  et~al.}]{chen2019mmdetection}
Chen, K.; Wang, J.; Pang, J.; Cao, Y.; Xiong, Y.; Li, X.; Sun, S.; Feng, W.;
  Liu, Z.; Xu, J.; et~al. 2019.
\newblock MMDetection: Open mmlab detection toolbox and benchmark.
\newblock \emph{arXiv preprint arXiv:1906.07155}.

\bibitem[{Fu et~al.(2019)Fu, Liu, Tian, Li, Bao, Fang, and Lu}]{fu2019dual}
Fu, J.; Liu, J.; Tian, H.; Li, Y.; Bao, Y.; Fang, Z.; and Lu, H. 2019.
\newblock Dual attention network for scene segmentation.
\newblock In \emph{Proceedings of the IEEE/CVF Conference on Computer Vision
  and Pattern Recognition}, 3146--3154.

\bibitem[{Gao et~al.(2019)Gao, Xie, Wang, and Li}]{gao2019global}
Gao, Z.; Xie, J.; Wang, Q.; and Li, P. 2019.
\newblock Global second-order pooling convolutional networks.
\newblock In \emph{Proceedings of the IEEE/CVF Conference on Computer Vision
  and Pattern Recognition}, 3024--3033.

\bibitem[{Gregorutti, Michel, and
  Saint-Pierre(2017)}]{gregorutti2017correlation}
Gregorutti, B.; Michel, B.; and Saint-Pierre, P. 2017.
\newblock Correlation and variable importance in random forests.
\newblock \emph{Statistics and Computing}, 27(3): 659--678.

\bibitem[{He et~al.(2017)He, Gkioxari, Doll{\'a}r, and Girshick}]{he2017mask}
He, K.; Gkioxari, G.; Doll{\'a}r, P.; and Girshick, R. 2017.
\newblock Mask r-cnn.
\newblock In \emph{Proceedings of the IEEE international conference on computer
  vision}, 2961--2969.

\bibitem[{He et~al.(2016)He, Zhang, Ren, and Sun}]{he2016deep}
He, K.; Zhang, X.; Ren, S.; and Sun, J. 2016.
\newblock Deep residual learning for image recognition.
\newblock In \emph{Proceedings of the IEEE conference on computer vision and
  pattern recognition}, 770--778.

\bibitem[{He et~al.(2019)He, Zhang, Zhang, Zhang, Xie, and Li}]{he2019bag}
He, T.; Zhang, Z.; Zhang, H.; Zhang, Z.; Xie, J.; and Li, M. 2019.
\newblock Bag of tricks for image classification with convolutional neural
  networks.
\newblock In \emph{Proceedings of the IEEE/CVF Conference on Computer Vision
  and Pattern Recognition}, 558--567.

\bibitem[{Hochreiter and Schmidhuber(1997)}]{hochreiter1997long}
Hochreiter, S.; and Schmidhuber, J. 1997.
\newblock Long short-term memory.
\newblock \emph{Neural computation}, 9(8): 1735--1780.

\bibitem[{Hu, Shen, and Sun(2018)}]{hu2018squeeze}
Hu, J.; Shen, L.; and Sun, G. 2018.
\newblock Squeeze-and-excitation networks.
\newblock In \emph{Proceedings of the IEEE conference on computer vision and
  pattern recognition}, 7132--7141.

\bibitem[{Huang et~al.(2017)Huang, Liu, Van Der~Maaten, and
  Weinberger}]{huang2017densely}
Huang, G.; Liu, Z.; Van Der~Maaten, L.; and Weinberger, K.~Q. 2017.
\newblock Densely connected convolutional networks.
\newblock In \emph{Proceedings of the IEEE conference on computer vision and
  pattern recognition}, 4700--4708.

\bibitem[{Huang et~al.(2020)Huang, Liang, Liang, and Yang}]{huang2020dianet}
Huang, Z.; Liang, S.; Liang, M.; and Yang, H. 2020.
\newblock Dianet: Dense-and-implicit attention network.
\newblock In \emph{Proceedings of the AAAI Conference on Artificial
  Intelligence}, volume~34, 4206--4214.

\bibitem[{Krizhevsky, Sutskever, and Hinton(2012)}]{krizhevsky2012imagenet}
Krizhevsky, A.; Sutskever, I.; and Hinton, G.~E. 2012.
\newblock Imagenet classification with deep convolutional neural networks.
\newblock \emph{Advances in neural information processing systems}, 25:
  1097--1105.

\bibitem[{Li and Chen(2020)}]{li2020deep}
Li, D.; and Chen, Q. 2020.
\newblock Deep Reinforced Attention Learning for Quality-Aware Visual
  Recognition.
\newblock In \emph{European Conference on Computer Vision}, 493--509. Springer.

\bibitem[{Li et~al.(2019)Li, Wang, Hu, and Yang}]{li2019selective}
Li, X.; Wang, W.; Hu, X.; and Yang, J. 2019.
\newblock Selective kernel networks.
\newblock In \emph{Proceedings of the IEEE/CVF Conference on Computer Vision
  and Pattern Recognition}, 510--519.

\bibitem[{Lin et~al.(2014)Lin, Maire, Belongie, Hays, Perona, Ramanan,
  Doll{\'a}r, and Zitnick}]{lin2014microsoft}
Lin, T.-Y.; Maire, M.; Belongie, S.; Hays, J.; Perona, P.; Ramanan, D.;
  Doll{\'a}r, P.; and Zitnick, C.~L. 2014.
\newblock Microsoft coco: Common objects in context.
\newblock In \emph{European conference on computer vision}, 740--755. Springer.

\bibitem[{Nair and Hinton(2010)}]{nair2010rectified}
Nair, V.; and Hinton, G.~E. 2010.
\newblock Rectified linear units improve restricted boltzmann machines.
\newblock In \emph{Icml}.

\bibitem[{Park et~al.(2018)Park, Woo, Lee, and Kweon}]{park2018bam}
Park, J.; Woo, S.; Lee, J.-Y.; and Kweon, I.~S. 2018.
\newblock Bam: Bottleneck attention module.
\newblock \emph{arXiv preprint arXiv:1807.06514}.

\bibitem[{Qin et~al.(2020)Qin, Zhang, Wu, and Li}]{qin2020fcanet}
Qin, Z.; Zhang, P.; Wu, F.; and Li, X. 2020.
\newblock Fcanet: Frequency channel attention networks.
\newblock \emph{arXiv preprint arXiv:2012.11879}.

\bibitem[{Ren et~al.(2016)Ren, He, Girshick, and Sun}]{ren2016faster}
Ren, S.; He, K.; Girshick, R.; and Sun, J. 2016.
\newblock Faster R-CNN: towards real-time object detection with region proposal
  networks.
\newblock \emph{IEEE transactions on pattern analysis and machine
  intelligence}, 39(6): 1137--1149.

\bibitem[{Ronneberger, Fischer, and Brox(2015)}]{ronneberger2015u}
Ronneberger, O.; Fischer, P.; and Brox, T. 2015.
\newblock U-net: Convolutional networks for biomedical image segmentation.
\newblock In \emph{International Conference on Medical image computing and
  computer-assisted intervention}, 234--241. Springer.

\bibitem[{Russakovsky et~al.(2015)Russakovsky, Deng, Su, Krause, Satheesh, Ma,
  Huang, Karpathy, Khosla, Bernstein et~al.}]{russakovsky2015imagenet}
Russakovsky, O.; Deng, J.; Su, H.; Krause, J.; Satheesh, S.; Ma, S.; Huang, Z.;
  Karpathy, A.; Khosla, A.; Bernstein, M.; et~al. 2015.
\newblock Imagenet large scale visual recognition challenge.
\newblock \emph{International journal of computer vision}, 115(3): 211--252.

\bibitem[{Simonyan and Zisserman(2014)}]{simonyan2014very}
Simonyan, K.; and Zisserman, A. 2014.
\newblock Very deep convolutional networks for large-scale image recognition.
\newblock \emph{arXiv preprint arXiv:1409.1556}.

\bibitem[{Wang et~al.(2020)Wang, Wu, Zhu, Li, Zuo, and Hu}]{wang2020eca}
Wang, Q.; Wu, B.; Zhu, P.; Li, P.; Zuo, W.; and Hu, Q. 2020.
\newblock ECA-Net: efficient channel attention for deep convolutional neural
  networks, 2020 IEEE.
\newblock In \emph{CVF Conference on Computer Vision and Pattern Recognition
  (CVPR). IEEE}.

\bibitem[{Wang et~al.(2018)Wang, Girshick, Gupta, and He}]{wang2018non}
Wang, X.; Girshick, R.; Gupta, A.; and He, K. 2018.
\newblock Non-local neural networks.
\newblock In \emph{Proceedings of the IEEE conference on computer vision and
  pattern recognition}, 7794--7803.

\bibitem[{Wang et~al.(2021)Wang, Yang, Bai, Zhang, Bai, Yu, Zhang, Huang, and
  Tong}]{wang2021evolving}
Wang, Y.; Yang, Y.; Bai, J.; Zhang, M.; Bai, J.; Yu, J.; Zhang, C.; Huang, G.;
  and Tong, Y. 2021.
\newblock Evolving attention with residual convolutions.
\newblock \emph{arXiv preprint arXiv:2102.12895}.

\bibitem[{Woo et~al.(2018)Woo, Park, Lee, and Kweon}]{woo2018cbam}
Woo, S.; Park, J.; Lee, J.-Y.; and Kweon, I.~S. 2018.
\newblock Cbam: Convolutional block attention module.
\newblock In \emph{Proceedings of the European conference on computer vision
  (ECCV)}, 3--19.

\bibitem[{Xu et~al.(2015)Xu, Ba, Kiros, Cho, Courville, Salakhudinov, Zemel,
  and Bengio}]{xu2015show}
Xu, K.; Ba, J.; Kiros, R.; Cho, K.; Courville, A.; Salakhudinov, R.; Zemel, R.;
  and Bengio, Y. 2015.
\newblock Show, attend and tell: Neural image caption generation with visual
  attention.
\newblock In \emph{International conference on machine learning}, 2048--2057.
  PMLR.

\bibitem[{Zhang et~al.(2020)Zhang, Wu, Zhang, Zhu, Lin, Zhang, Sun, He,
  Mueller, Manmatha et~al.}]{zhang2020resnest}
Zhang, H.; Wu, C.; Zhang, Z.; Zhu, Y.; Lin, H.; Zhang, Z.; Sun, Y.; He, T.;
  Mueller, J.; Manmatha, R.; et~al. 2020.
\newblock Resnest: Split-attention networks.
\newblock \emph{arXiv preprint arXiv:2004.08955}.

\bibitem[{Zhang and Yang(2021)}]{zhang2021sa}
Zhang, Q.-L.; and Yang, Y.-B. 2021.
\newblock Sa-net: Shuffle attention for deep convolutional neural networks.
\newblock In \emph{ICASSP 2021-2021 IEEE International Conference on Acoustics,
  Speech and Signal Processing (ICASSP)}, 2235--2239. IEEE.

\end{thebibliography}
\end{document}